\documentclass[conference]{IEEEtran}
\IEEEoverridecommandlockouts
\usepackage{cite}
\usepackage{amsmath,amssymb,amsfonts}
\usepackage{algorithm}
\usepackage{algorithmic}
\usepackage{graphicx}
\usepackage{textcomp}
\usepackage{xcolor}
\usepackage[T1]{fontenc}
\usepackage[utf8]{inputenc}

\def\BibTeX{{\rm B\kern-.05em{\sc i\kern-.025em b}\kern-.08em
    T\kern-.1667em\lower.7ex\hbox{E}\kern-.125emX}}

\usepackage{tikz}
\usetikzlibrary{arrows.meta,positioning,fit}

\begin{document}

{\title{A Smart-Scheduled Hybrid EKF–FGO State Estimation}
}

\author{
\IEEEauthorblockN{Eric Levy}
\IEEEauthorblockA{
\textit{Electrical, Computer, and Biomedical Engineering}\\
\textit{ Toronto Metropolitan University} \\
Toronto, ON, Canada \\
eric.j.levy@torontomu.ca\\
0009-0004-2724-7668
}
\and
\IEEEauthorblockN{Soosan Beheshti}
\IEEEauthorblockA{
\textit{Electrical, Computer, and Biomedical Engineering} \\
\textit{Toronto Metropolitan University} \\
Toronto, ON, Canada \\
soosan@torontomu.ca\\
0000-0001-7161-5887
}
}

\maketitle

\begin{abstract}
Reliable state estimation in robotics requires balancing estimation accuracy with computational cost. Filtering-based methods such as the Extended Kalman Filter (EKF) provide efficient real-time updates but accumulate drift due to linearisation. In contrast, optimisation-based approaches using factor graph optimisation (FGO) improve global consistency at higher computational cost.
This paper investigates optimisation scheduling as an explicit design variable in hybrid estimation systems. We introduce a scheduled hybrid EKF–FGO framework in which recursive filtering is performed at every time step, while batch optimisation is invoked periodically. By fixing the solver structure and effort, the framework enables controlled analysis of scheduling effects.
Simulation results in a planar SLAM setting show that scheduling strongly influences pre-optimisation drift and computational cost. In particular, we observe an asymmetric trade-off: drift increases gradually with longer scheduling intervals, while computational cost decreases rapidly. This leads to operating regimes in which most of the benefits of global optimisation are retained at a fraction of the cost.
These findings highlight optimisation scheduling as a key and under explored design dimension in hybrid state estimation systems.

\end{abstract}


\section{Introduction}

State estimation under uncertainty requires inferring system states from noisy measurements while operating under limited computational resources. In many estimation pipelines, this creates a fundamental trade-off between estimation accuracy and computational cost, particularly when combining fast recursive updates with more expensive global optimisation.

Simultaneous Localisation and Mapping (SLAM) provides a representative setting in which this trade-off is clearly exposed. Filtering-based approaches such as the Extended Kalman filter (EKF) enable efficient online estimation but accumulate drift due to linearisation error. Optimisation-based approaches using factor graph optimization (FGO) improve global consistency by jointly refining multiple states, at the cost of increased computational demand.

While these characteristics are well understood, the temporal organisation of estimation and optimisation—specifically, when global optimisation should be invoked—has received comparatively little systematic analysis. In practice, optimisation scheduling is often treated as an implicit design choice rather than an explicit variable that can be analysed and tuned.

Hybrid estimation strategies that combine filtering and optimisation offer a natural framework for exploring this question. By interleaving fast recursive estimation with periodic batch optimisation, such approaches expose scheduling as a tunable design parameter rather than a fixed implementation choice. However, the effect of scheduling on estimator behaviour, transient error dynamics, and computational cost remains insufficiently characterised.

This work investigates optimisation scheduling as an explicit design variable in hybrid state estimation systems. To facilitate this study, we introduce a scheduled hybrid EKF--FGO estimation framework in which recursive filtering and batch optimisation are coupled through explicit bidirectional information exchange and controlled optimisation scheduling. While motivated by SLAM, the analysis is not specific to SLAM itself; rather, SLAM provides a familiar context in which scheduling effects are readily observable. 

Through simulation experiments, we show that scheduling plays a central role in shaping estimator behaviour, revealing a structured trade-off between intermediate estimation accuracy and computational cost. The primary contribution of this work is the experimental characterisation of optimisation scheduling as a mechanism for navigating this trade space between filtering and batch optimisation.

\section{Background and Motivation}

Early work in state estimation and SLAM established filtering-based approaches such as the EKF as a practical solution for recursive estimation from noisy sensor data \cite{b1,b2,b3,b4,b5,b6}. EKF-based methods are attractive due to their relatively low computational cost and low-latency state estimates, but reliance on first-order linearisation introduces well-known limitations, including estimator inconsistency and the accumulation of drift over time \cite{b5,b9,b10}.

Optimisation-based formulations address many of these limitations by casting state estimation as a maximum a posteriori (MAP) inference problem over a factor graph, solved via nonlinear least-squares optimisation. Such approaches have been widely adopted in both batch and incremental SLAM systems \cite{b7,b8,b11,b12,b13,b14}, and achieve improved global consistency and reduced drift compared to filtering-based methods. These benefits come at the cost of increased computational demand, particularly when large windows of past states are repeatedly re-optimised \cite{b11}.

Incremental methods such as iSAM2 \cite{b8} address computational efficiency through architectural innovations including variable reordering and Bayes tree factorization, but represent a different point in the design space than the scheduling question examined here. The present work treats scheduling as an orthogonal concern: given any batch optimisation capability, how frequently should it be invoked, independent of solver architecture. Hybrid and sliding-window estimation strategies combine filtering-based propagation with intermittent optimisation to balance real-time performance with global consistency \cite{b15,b16,b17,b18,b19,b20}. Sliding-window estimators maintain a bounded horizon of recent states while marginalising older variables to control computational complexity \cite{b16,b17}. In many such systems, optimisation scheduling and filter–optimiser interaction are treated implicitly as fixed design choices rather than explicit objects of study. As a result, the effect of optimisation scheduling on estimator behaviour, including drift correction, transient error dynamics, and computational cost, remains underexplored.

These observations motivate the approach taken in this work. Rather than proposing a new SLAM algorithm, we introduce a hybrid EKF--FGO framework designed to make the interaction between filtering and optimisation explicit and experimentally controllable. By injecting the EKF belief as a Gaussian prior factor within the optimisation problem and treating the optimisation schedule as an independent design variable, the proposed framework enables systematic investigation of scheduling effects. SLAM serves as a motivating application, but the questions addressed here arise more broadly in hybrid state estimation systems.

\section{Smart Scheduled Hybrid EKF--FGO Estimation Framework}

This section presents the Smart Scheduled Hybrid (SSH) EKF--FGO estimation framework introduced in this work. The framework is designed to make the interaction between recursive filtering and batch optimisation explicit and experimentally controllable, with particular emphasis on optimisation scheduling.

\subsection{EKF Belief Representation}
At each time step $t$, the EKF maintains a Gaussian belief over the system state, summarised by a mean vector $\mu_t$ and a covariance matrix $P_t$, representing the estimated state and its associated uncertainty. The EKF belief arises from standard prediction and measurement update steps applied to a nonlinear state--space model with Gaussian noise. In this work, $\mu_t$ and $P_t$ are treated as sufficient statistics of the EKF belief and are explicitly reused within the proposed hybrid estimation framework.

\subsection{Factor Graph Formulation}
In the factor graph formulation, state estimation is cast as a maximum a posteriori (MAP) inference problem over a set of state variables. Let
\begin{align}
    \mathcal{X} = \{x_0, x_1, \ldots, x_T, \ell_1, \ldots, \ell_M\}
\end{align}
denote the collection of robot poses over a finite time horizon $T$ and a map of $M$ landmarks. Probabilistic constraints on these variables are represented by factors encoding motion, measurement, or prior information.

The MAP estimate of the global state is obtained by minimising the sum of squared residuals induced by all factors,
\begin{align}
    \mathcal{X}^* = \arg \min_{\mathcal{X}} \sum_i \|\rho_i(\mathcal{X})\|^2_{\Sigma_i^{-1}},
\end{align}
where each residual $\rho_i(\cdot)$ corresponds to a factor with covariance $\Sigma_i$. This formulation is retained unchanged in the proposed hybrid framework, allowing the effects of scheduling and estimator interaction to be examined independently of solver design.

\subsection{Smart Scheduled Hybrid (SSH) EKF--FGO Framework}
The proposed SSH EKF--FGO framework combines recursive filtering and batch optimisation through explicit estimator scheduling. The EKF provides efficient step-by-step propagation of a local estimate, while FGO is invoked periodically to enforce global consistency. Importantly, the SSH framework modifies when optimisation is performed and how it is initialised, without altering the underlying optimisation algorithm.

\subsubsection{Scheduling Strategy}
Let $W$ denote the optimisation scheduling interval. The EKF runs at every time step, producing a rolling estimate $(\mu_t, P_t)$. Full graph optimisation is triggered only at discrete time steps
\[
t \in \{W, 2W, 3W, \ldots\},
\]
with intermediate state estimates propagated exclusively by the EKF. At each optimisation event, a factor graph is constructed over the current optimisation window and augmented with an EKF-derived prior. In this study, the scheduling interval $W$ is fixed for each experiment to isolate scheduling effects; adaptive or state-dependent scheduling strategies are left for future work.

\subsubsection{EKF Prior Residual and Factor}
At each optimisation event, the EKF estimate is incorporated into the factor graph as a unary prior on the current pose variable $x_t$,
\begin{align}
    \rho^{\text{prior}}(x_t) = x_t - \mu_t,
\end{align}
with corresponding factor
\begin{align}
    \varphi^{\text{prior}}(x_t) \triangleq
    \rho^{\text{prior}}(x_t)^\top (\alpha P_t)^{-1} \rho^{\text{prior}}(x_t),
\end{align}
where $P_t$ is the EKF covariance matrix and $\alpha > 0$ controls the strength of the prior. Smaller values of $\alpha$ enforce stronger adherence to the EKF estimate, while larger values recover behaviour closer to classical batch FGO.

\subsubsection{Hybrid Optimisation Problem}
With the EKF prior included, the hybrid optimisation problem takes the form
\begin{align}
    \mathcal{X}^* =
    \arg \min_{\mathcal{X}}
    \left(
        \sum_i \varphi_i(\mathcal{X}) +
        \varphi^{\text{prior}}(x_t)
    \right),
\end{align}
where $\varphi_i(\mathcal{X})$ denotes the motion and measurement factors defined in the factor graph formulation. The resulting objective remains a nonlinear least-squares problem and is solved using the Gauss--Newton (GN) method for a fixed number of iterations $K$.

\subsubsection{Interaction Between EKF and FGO}
Following optimisation, the EKF is reinitialised using the optimised state to maintain consistency between filtering and optimisation. From a scheduling perspective, the framework exposes two mechanisms governing estimator behaviour: the scheduling interval $W$, which determines how frequently global corrections occur, and EKF propagation, which governs error accumulation between optimisation events. These interactions form the basis of the experimental analysis presented in the results section.

\begin{algorithm}[t]
\caption{Smart Scheduled Hybrid (SSH) EKF--FGO Scheduling Framework}
\label{alg:hybrid_ekf_fgo}
\begin{algorithmic}[1]

\STATE \textbf{Input:} Initial state estimate $\hat{x}_0$, covariance $P_0$
\STATE \textbf{Input:} Control inputs $\{u_t\}$, observations $\{z_t\}$
\STATE \textbf{Input:} Sliding window length $W$, GN iterations $K$

\FOR{$t = 1$ to $T$}

    \STATE \textbf{EKF Prediction:}
    \STATE \hspace{0.5cm} $\hat{x}_{t|t-1} \leftarrow f(\hat{x}_{t-1}, u_t)$
    \STATE \hspace{0.5cm} $P_{t|t-1} \leftarrow F_t P_{t-1} F_t^\top + Q_t$

    \STATE \textbf{EKF Update:}
    \STATE \hspace{0.5cm} $\hat{x}_t \leftarrow \hat{x}_{t|t-1} + K_t (z_t - h(\hat{x}_{t|t-1}))$
    \STATE \hspace{0.5cm} $P_t \leftarrow (I - K_t H_t) P_{t|t-1}$

    \IF{$t \bmod W = 0$}

        \STATE Construct factor graph over window $[t-W+1, t]$
        \STATE Add motion and measurement factors
        \STATE Add EKF prior factor $(\hat{x}_t, P_t)$
        \STATE Run Gauss--Newton optimisation for $K$ iterations
        \STATE Extract optimised state estimates

    \ENDIF

\ENDFOR

\STATE \textbf{Output:} Estimated state trajectory $\{\hat{x}_t\}$

\end{algorithmic}
\end{algorithm}

\section{Experimental Setup}
\begin{figure*}[t]
  \centering
  \includegraphics[width=\linewidth]{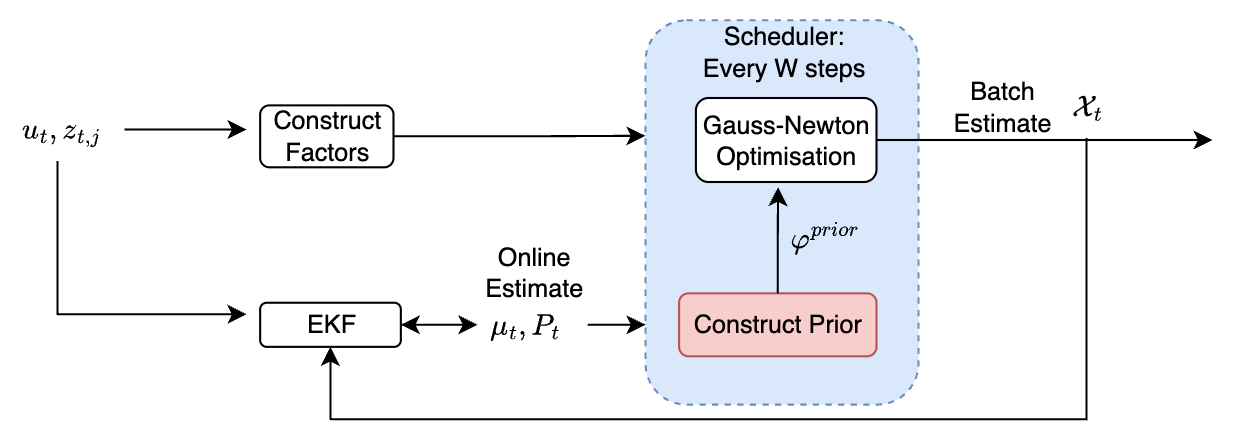}
  \caption{Block diagram of the hybrid EKF--FGO estimator. Control inputs $u_t$ and landmark observations $z_{t,j}$ are processed at every time step by the EKF to produce a rolling state estimate $(\mu_t, P_t)$ used for online pose estimation. A scheduler triggers factor graph construction and FGO optimization every $W$ steps, using the same motion and measurement data together with an EKF-derived prior $(\mu_t, \alpha P_t)$. The resulting batch estimate $\mathcal{X}^*$ is used to reinitialize the EKF, providing feedback between filtering and batch optimization.}
  \label{fig:hybrid_block_diagram}
\end{figure*}

This section describes the experimental configuration used to evaluate the proposed hybrid EKF-FGO scheduling framework. All experiments are conducted in simulation to allow controlled variation of scheduling parameters and direct comparison between estimation methods under identical conditions.

\subsection{Simulation Environment}\label{Simulation Environment}
We consider a planar SLAM scenario in which a robot moves through a static environment containing a fixed set of landmarks. The robot follows a predefined smooth circular trajectory with constant curvature and radius $R$. In all experiments, the trajectory radius is fixed at $R = 1$~m, resulting in a motion profile with uniform turning rate. This choice provides a controlled setting for analysing scheduling effects under consistent geometric conditions. Since curvature directly influences error accumulation during propagation, the reported results correspond to this specific motion regime; evaluating the effect of varying trajectory curvature is left for future work.

\subsection{Estimators and Parameters}\label{Estimators and Parameters}
Three estimators are evaluated:
\begin{enumerate}
    \item EKF, which performs recursive joint estimation of robot poses and landmark positions.
    \item Classical Factor Graph Optimisation, which performs periodic batch optimisation over the SLAM state without filtering-based propagation between optimisation events.
    \item SSH EKF-FGO, which combines EKF-based propagation with periodically scheduled graph optimisation using an EKF-derived prior.
\end{enumerate}
All the estimators use the same motion model, observation model, and noise covariances. The EKF prior scaling parameter $\alpha$ was held fixed at 0.5 for all experiments; sensitivity to $\alpha$ is left for future work. For FGO-based methods, the GN solver is run for a fixed number of iterations $K$ in all cases. This constraint ensures that observed differences in estimation accuracy and computational cost arise from scheduling and estimator structure rather than differences in solver effort.

\subsection{Scheduling Configuration}\label{Scheduling Configuration}
For the classical FGO and hybrid EKF-FGO methods, the optimisation is executed periodically according to a scheduling interval $W$ optimisation is triggered at time steps $t\in \{W, 2W, 3W,...\}$. Between optimisation events the EKF propagates state estimates at every time step for the EKF and hybrid methods, where the classical FGO baseline propagates state estimates using the motion model alone. In the hybrid framework, the EKF estimate $(\mu_t, P_t)$ at each scheduling time is incorporated into the factor graph as a prior factor, with covariance scaled by the fixed parameter $\alpha$. To study the effect of estimator scheduling, experiments are conducted over a range of scheduling intervals $W$, while holding the solver iteration count $K$ fixed. In all experiments reported in this paper, the EKF is reinitialised with the optimised pose estimate at each scheduling event, ensuring consistency between the filtering and optimisation components.

\subsection{Evaluation Metrics}\label{Evaluation Metrics}
Estimator performance is evaluated using metrics designed to capture both accuracy and scheduling effects. To quantify estimation accuracy, we measure pose error with respect to ground truth over the duration of each experiment. In addition, to explicitly characterise the effect of estimator scheduling, we introduce a pre-optimisation drift metric, defined as the pose error accumulated immediately prior to each optimisation event. This metric reflects how far an estimator is allowed to drift before a global correction is applied and provides a direct insight into the interaction between propagation strategy and optimisation frequency. Pre-optimisation drift is reported alongside post-optimisation accuracy and computational cost to provide a comprehensive view of estimator behaviour under different scheduling configurations.

\section{Results}\label{results}

\begin{figure}[t]
    \centering
    \includegraphics[width=\columnwidth]{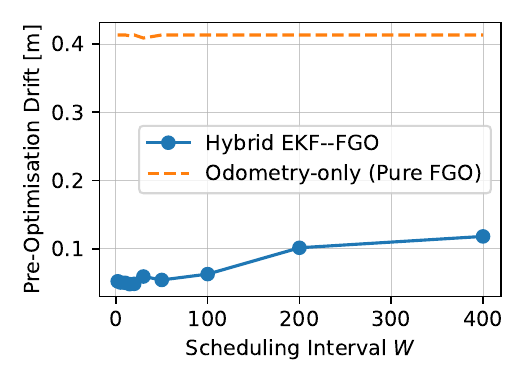}
    \caption{Average pre-optimisation drift as a function of scheduling interval $W$ for the circular trajectory. The odometry-only curve corresponds to a classical batch FGO configuration in which state propagation is performed using the motion model alone, with a single batch optimisation at the final time step. The hybrid EKF--FGO estimator consistently achieves substantially lower drift across all scheduling intervals.}
    \label{fig:circle_preopt_drift}
\end{figure}

\begin{figure}[t]
  \centering
  \includegraphics[width=\columnwidth]{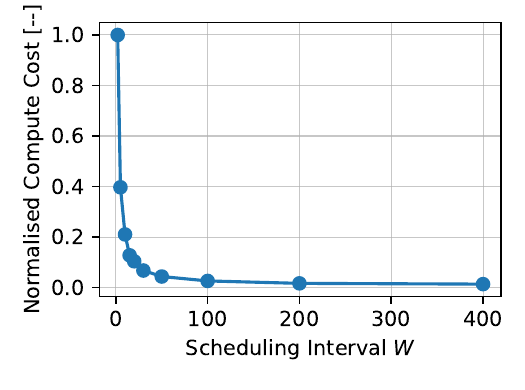}
  \caption{Normalized computational cost per time step for the hybrid EKF--FGO estimator as a function of scheduling interval $W$. Cost is measured as average wall-clock optimisation time per step and normalised by the slowest configuration, yielding a dimensionless quantity independent of hardware.}
  \label{fig:circle_norm_cost}
\end{figure}

\begin{table*}[t]
\centering
\caption{Pre-optimisation drift, drift reduction, and normalised computational cost for circular trajectories under varying scheduling intervals W.}
\label{tab:circle_preopt_results}
\begin{tabular}{rccccccc}
\hline
$W$ &
Hybrid Pre-Opt &
Odom Pre-Opt &
Drift Reduction &
Norm. Compute &
ATE Hybrid &
ATE Odom \\
&
Drift [m] &
Drift [m] &
[\%] &
Cost [--] &
[m] &
[m] \\
\hline
   2   & 0.052 & 0.413 & 87.3 & 1.000 & 0.050 & 0.484 \\
   5   & 0.051 & 0.413 & 87.8 & 0.345 & 0.050 & 0.484 \\
  10   & 0.050 & 0.413 & 87.9 & 0.171 & 0.050 & 0.484 \\
  15   & 0.048 & 0.412 & 88.3 & 0.111 & 0.052 & 0.484 \\
  20   & 0.049 & 0.413 & 88.3 & 0.087 & 0.050 & 0.484 \\
  30   & 0.059 & 0.409 & 85.5 & 0.051 & 0.052 & 0.484 \\
  50   & 0.054 & 0.413 & 86.9 & 0.039 & 0.050 & 0.484 \\
 100   & 0.063 & 0.413 & 84.8 & 0.027 & 0.050 & 0.484 \\
 200   & 0.101 & 0.413 & 75.5 & 0.017 & 0.050 & 0.484 \\
 400   & 0.118 & 0.413 & 71.4 & 0.008 & 0.050 & 0.484 \\
\hline
\end{tabular}
\end{table*}

This section evaluates the proposed EKF-FGO hybrid estimator under varying optimisation scheduling intervals using a circular trajectory. All experiments are conducted with identical system models, noise parameter, and solver iteration limits. Performance differences therefore arise solely from estimator structure and scheduling. The primary focus of this evaluation is pre-optimisation drift, which quantifies the amount of pose error accumulated between optimisation events and directly reflects intermediate estimator behaviour. This metric enables direct comparison between the hybrid estimator and a batch baseline that performs a single optimisation at the end of the trajectory.

\subsection{Pre-Optimisation Drift Reduction}
Figure \ref{fig:circle_preopt_drift} compares average pre-optimisation drift as a function of scheduling interval $W$ for the hybrid estimator and the odometry-only baseline. The odometry curve corresponds to a classical batch FGO formulation in which the state is propagated without filtering and optimised only once at the final time step. Consequently, its pre-optimisation drift represents the total error accumulated prior to that final batch optimisation and is effectively independent of $W$. Across all tested scheduling intervals, the hybrid estimator exhibits substantially lower pre-optimisation drift than the odometry-only baseline. For scheduling intervals up to $W=50$, hybrid drift remains below approximately 0.06 m, corresponding to an 80-90\% reduction relative to the odometry baseline. Even for very infrequent optimisation, such as $W=200$, hybrid drift remains three to four times lower than odometry-only propagation. These results demonstrate that EKF-based propagation provides a strong constraint on error growth between optimisation events, significantly reducing intermediate drift even when global optimisation is performed rarely.

\subsection{Effects of Scheduling Interval}
As the scheduling interval increases, hybrid pre-optimisation drift increases gradually, reflecting longer periods of EKF-only propagation between optimisation events. Importantly, this increase is smooth and monotonic, indicating stable interaction between filtering and batch optimisation under the corrected implementation. While larger values of $W$ permit greater drift accumulation, the hybrid estimator consistently maintains drift well below the odometry-only baseline for all tested schedules. This behaviour highlights the robustness of the hybrid approach across a wide range of scheduling intervals and confirms that the EKF provides meaningful intermediate accuracy even when optimisation frequency is significantly reduced.

This behaviour can be interpreted through the differing roles of filtering and optimisation in the hybrid framework. EKF-based propagation provides locally consistent updates that limit the rate of error growth between optimisation events, leading to gradual accumulation of drift as the scheduling interval increases. In contrast, the computational cost of batch optimisation depends on both the number of variables in the optimisation window and the frequency of optimisation. As the scheduling interval increases, optimisation is invoked less frequently, resulting in a rapid reduction in computational cost. This difference in scaling explains the observed asymmetry: drift increases gradually with scheduling interval, while computational cost decreases rapidly.

\subsection{Final Trajectory Accuracy}
Table \ref{tab:circle_preopt_results} reports final absolute trajectory error (ATE) for all estimators. Final ATE for the hybrid estimator remains nearly constant across all scheduling intervals and closely matches the accuracy of classical batch FGO. In contrast, odometry-only propagation results in substantially higher final error.

These results indicate that scheduling primarily affects intermediate estimation behaviour rather than final accuracy, provided that at least one batch optimisation is performed. The hybrid framework therefore achieves EKF-level stability during propagation while retaining FGO-level accuracy after optimisation.

\subsection{Computation Implications}
Increasing the scheduling interval reduces the frequency of batch optimisation and leads to substantial reductions in computational cost. When combined with the observed drift behaviour, the results reveal a favourable operating regime in which most of the drift reduction benefit of frequent optimisation is retained at a fraction of the computational cost. In particular, moderate scheduling intervals (e.g. $W \in [10, 50]$) achieve pre-optimisation drift reductions exceeding 85\% relative to odometry-only propagation, while allowing optimisation frequency and corresponding computational load to be reduced by more than an order of magnitude. This behaviour arises from a strongly asymmetric trade-off: pre-optimisation drift grows gradually with increasing $W$, whereas computational cost decreases rapidly, resulting in a wide range of scheduling intervals that offer favourable accuracy--efficiency balance.

\subsection{Summary of Results}
These results demonstrate that hybrid EKF-FGO scheduling provides effective drift reduction (up to 90\%) relative to odometry-only propagation while enabling substantial computational savings, with performance degrading smoothly as optimization frequency decreases. Final trajectory accuracy remains comparable to pure FGO and largely independent of scheduling interval. The findings confirm that scheduling provides a practical and effective mechanism for balancing estimation accuracy and computational efficiency, exposing a design dimension that can be tuned to application-specific performance requirements.

\section{Conclusion}
In this paper we examined the effects of estimator scheduling on accuracy and computational cost in batch-based state estimation systems motivated by SLAM backends. Rather than proposing a new inference algorithm, the work focused on experimentally characterising how the frequency of batch optimisation influences the intermediate error growth and runtime, and on identifying regimes in which accuracy can be preserved while significantly reducing computational effort. To enable this analysis, the SSH EKF-FGO framework was used as a practical evaluation tool. In this setup, recursive filtering provides efficient state propagation between optimisation events, while periodic batch optimisation enforces global consistency. This hybrid structure is not intended to replace established incremental solvers, but rather to facilitate controlled experiments on scheduling behaviour and intermediate estimation performance. 

Experimental results on a circular trajectory show that EKF-based propagation substantially reduces pre-optimisation drift relative to an odometry-only baseline across all tested scheduling intervals, achieving reductions of approximately 70-90\%. At the same time, final trajectory accuracy remains comparable to pure batch optimisation, while computational cost decreases rapidly as optimisation frequency is reduced. These results demonstrate that scheduling has a pronounced effect on intermediate estimation behaviour, even when final accuracy is largely unchanged.

Overall, the findings highlight estimator scheduling as an important and often under-explored design dimension in batch-based state estimation systems. By explicitly evaluating the trade-off between intermediate accuracy and computational cost, this work provides practical insights into how optimisation frequency can be chosen to balance performance and efficiency. Future work will extend this analysis to more complex trajectories, longer horizons, and adaptive scheduling strategies that respond to uncertainty or motion dynamics.

\end{document}